# Sparse Stochastic Finite-State Controllers for POMDPs

**Eric A. Hansen**
Dept. of Computer Science and Engineering
Mississippi State University
Mississippi State, MS 39762
hansen@cse.msstate.edu

## Abstract

Bounded policy iteration is an approach to solving infinite-horizon POMDPs that represents policies as stochastic finite-state controllers and iteratively improves a controller by adjusting the parameters of each node using linear programming. In the original algorithm, the size of the linear programs, and thus the complexity of policy improvement, depends on the number of parameters of each node, which grows with the size of the controller. But in practice, the number of parameters of a node with non-zero values is often very small, and does not grow with the size of the controller. Based on this observation, we develop a version of bounded policy iteration that leverages the sparse structure of a stochastic finite-state controller. In each iteration, it improves a policy by the same amount as the original algorithm, but with much better scalability.

## 1 Introduction

Partially observable Markov decision processes (POMDPs) provide a framework for decision-theoretic planning problems where actions need to be taken based on imperfect state information. Many researchers have shown that a policy for an infinite-horizon POMDP can be represented by a finite-state controller. In some cases, this is a deterministic controller in which a single action is associated with each node, and an observation results in a deterministic transition to a successor node (Kaelbling, Littman, & Cassandra, 1998; Hansen, 1998; Meuleau, Kim, Kaelbling, & Cassandra, 1999a). In other cases, it is a stochastic controller in which actions are selected based on a probability distribution associated with each node, and an observation results in a probabilistic transition to a successor node (Platzman, 1981; Meuleau, Peshkin, Kim, & Kaelbling, 1999b; Baxter & Bartlett, 2001; Poupart & Boutilier, 2004; Amato, Bernstein, & Zilberstein, 2007).

*Bounded policy iteration* (BPI) is an approach to solving infinite-horizon POMDPs that represents policies as stochastic finite-state controllers and iteratively improves a controller by adjusting the parameters of each node using linear programming, where the parameters specify the action selection and node transition probabilities of the node (Poupart & Boutilier, 2004). BPI is related to an exact policy iteration algorithm for POMDPs due to Hansen (1998), but differs from it by providing an elegant and effective approach to approximation in which bounding the size of the controller allows a tradeoff between planning time and plan quality. Originally developed as an approach to solving single-agent POMDPs, BPI has also been generalized for use in solving decentralized POMDPs (Bernstein, Hansen, & Zilberstein, 2005).

In BPI, the complexity of policy improvement depends on the size of the linear programs used to adjust the parameters of each node of the controller. In turn, this depends on the number of parameters of each node (as well as the size of the state space). In the original algorithm, each node of the controller has $|A|+|A||Z||\mathcal{N}|$ parameters, where $|A|$ in the number of actions, $|Z|$ is the number of observations, and $|\mathcal{N}|$ is the number of nodes of the controller. This assumes a fully-connected stochastic controller. In practice, however, most of these parameters have zero probabilities, and the number of parameters of a node with non-zero probabilities remains relatively constant as the number of nodes of the controller increases. Based on this observation, we propose a modified version of BPI that leverages a sparse representation of a stochastic finite-state controller. In each iteration, it improves the controller by the same amount as the original algorithm. But it does so by solving much smaller linear programs, where the number of variables in each linear program depends on the number of parameters of a node with non-zero values. Because the number of parameters of a node with non-zero values tends to remain relatively constant as the size of the controller grows, the complexity of each iteration of the modified version of BPI tends to grow only linearly with the number of nodes of a controller, which is a dramatic improvement in scalability compared to the original algorithm.

## 2 Background

We consider a discrete-time infinite-horizon POMDP with a finite set of states, $\mathcal{S}$, a finite set of actions, $\mathcal{A}$, and a finite set of observations, $\mathcal{Z}$. Each time period, the environment is in some state $s \in \mathcal{S}$, the agent takes an action $a \in \mathcal{A}$, the environment makes a transition to state $s' \in \mathcal{S}$ with probability $P(s'|s,a)$, and the agent observes $z \in Z$ with probability $P(z|s',a)$. In addition, the agent receives an immediate reward with expected value $R(s,a) \in \Re$. We assume the objective is to maximize expected total discounted reward, where $\beta \in (0,1]$ is the discount factor.

Since the state of the environment cannot be directly observed, we let $b$ denote an $|\mathcal{S}|$-dimensional vector of state probabilities, called a *belief state*, where $b(s)$ denotes the probability that the system is in state $s$. If action $a$ is taken and followed by observation $z$, the successor belief state, denoted $b_z^a$, is determined using Bayes' rule.

### 2.1 Policy representation and evaluation

A *policy* for a POMDP can be represented by a finite-state controller (FSC). A stochastic finite-state controller is a tuple $<\mathcal{N}, \psi, \eta>$ where $\mathcal{N}$ is a finite set of nodes, $\psi$ is an action selection function that specifies the probability, $\psi_n(a) = P(a|n)$, of selecting action $a \in A$ when the FSC is in node $n \in \mathcal{N}$, and $\eta$ is a node transition function that specifies the probability, $\eta_n(a,z,n') = P(n'|n,a,z)$, that the FSC will make a transition from node $n \in \mathcal{N}$ to node $n' \in \mathcal{N}$ after taking action $a \in A$ and receiving $z \in Z$.

The value function of a policy represented by a FSC is piecewise linear and convex, and can be computed exactly by solving the following system of linear equations, with one equation for each pair of node $n \in \mathcal{N}$ and state $s \in S$:

$$V_n(s) = \sum_{a \in A} \psi_n(a) R(s,a) + \qquad (1)$$
$$\beta \sum_{a,z,s,n'} \eta_n(a,z,n') P(s'|s,a) P(z|s',a) V_{n'}(s').$$

In this representation of the value function, there is one $|S|$-dimensional vector $V_n$ for each node $n \in \mathcal{N}$ of the controller. The value of any belief state $b$ is determined as follows,

$$V(b) = \max_{n \in N} \sum_{s \in S} b(s) V_n(s), \qquad (2)$$

and the controller is assumed to start in the node that maximizes the value of the initial belief state. The value function of an optimal policy satisfies the optimality equation,

$$V(b) = \max_{a \in A} \left\{ R(b,a) + \beta \sum_{z \in Z} P(z|b,a) V(b_z^a) \right\}, \qquad (3)$$

where $R(b,a) = \sum_{s \in S} b(s) R(s,a)$ and $P(z|b,a) = \sum_{s \in S} b(s) \sum_{s' \in S} P(s'|s,a) P(z|s',a)$.

---

Variables: $\epsilon$; $\psi_n(a), \forall a$; $\eta_n(a,z,n'), \forall a,z,n'$
Objective: Maximize $\epsilon$
Improvement constraints:
$V_n(s) + \epsilon \leq \sum_a \psi_n(a) R(s,a) +$
$\quad \gamma \sum_{a,z,n'} \eta_n(a,z,n') P(s'|s,a) P(z|s',a) V_{n'}(s'), \forall s$
Probability constraints:
$\sum_a \psi(a) = 1$
$\sum_{n'} \eta_n(a,z,n') = \psi(a), \forall a,z$
$\psi(a) \geq 0, \forall a$
$\eta_n(a,z,n') \geq 0, \forall a,z,n'$

Table 1: Linear program for improving a node $n$ of a stochastic finite-state controller.

### 2.2 Bounded policy iteration

Policy iteration algorithms iteratively improve a policy by alternating between two steps: policy evaluation and policy improvement. Hansen (1998) proposed a policy iteration algorithm for POMDPs that represents a policy as a deterministic finite-state controller. In the policy improvement step, it uses the dynamic programming update for POMDPs to add, merge and prune nodes of the controller. The algorithm is guaranteed to converge to an $\epsilon$-optimal controller and can detect convergence to an optimal policy.

A potential problem is that the number of nodes added in the policy improvement step can be large, and the controller can grow substantially in size from one iteration to the next. Because the complexity of the policy improvement step increases with the size of the controller, allowing the size of the controller to grow too fast can slow the rate of further improvement and limit scalability. The same problem occurs in value iteration, where the number of vectors in the piecewise linear and convex representation of the value function can grow too fast from one iteration to the next. Feng and Hansen (2001) describe a method for performing approximate dynamic programming updates that has the effect of reducing the number of vectors added to a value function in value iteration, or the number of nodes added to a controller in policy iteration.

For policy iteration, Poupart and Boutilier (2004) propose another approach that also has the benefit of avoiding an expensive dynamic programming update. Their *bounded policy iteration* algorithm controls growth in the size of the controller in two ways. First, a policy is represented as a stochastic finite-state controller that can be improved without increasing its size, by adjusting the action and node-transition probabilities of each node of the controller using linear programming. Second, when the controller cannot be improved further in this way, $k$ nodes are added to the controller, where $k$ is some small number greater than or equal to one. In the rest of this background section, we review each of these two steps in further detail.

**Algorithm 1** Bounded policy iteration
  **repeat**
    **repeat**
      {Policy evaluation}
      Solve the linear system given by Equation (1)
      {Policy improvement without adding nodes}
      **for** each node $n$ of the controller **do**
        solve the linear program in Table 1
        **if** $\epsilon > 0$ **then**
          update parameters and value vector of node
        **end if**
      **end for**
    **until** no improvement of controller
    {Policy improvement by adding nodes}
    find belief states reachable from tangent points
    create $k$ new nodes that improve their value
  **until** no new node is created

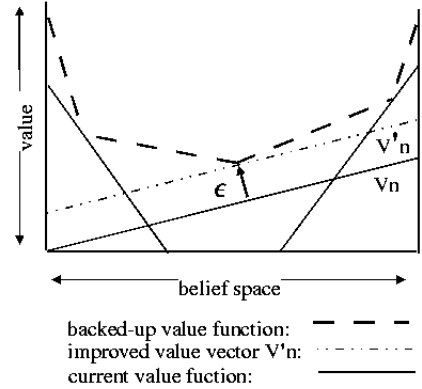

Figure 1: BPI can improve the value vector $V_n$ by an amount $\epsilon$ to obtain the improved value vector $V'_n$, which is tangent to the backed-up value function.

**Improving nodes** The first step attempts to improve a controller while keeping its size fixed. For each node $n$ of the controller, the linear program in Table 1 is solved. The linear program searches for action probabilities and node transition probabilities for the node that improve the value vector $V_n$ associated with the node by some amount $\epsilon$ for each state, where $\epsilon$ is the objective maximized by the linear program. If an improvement is found, the parameters of the node are updated accordingly. (The value vector may also be updated, and will be further improved during the policy evaluation step.)

Poupart and Boutilier (2004) show that the linear program can be interpreted as follows: it implicitly considers the vectors of the backed-up value function that would be created by performing a dynamic programming update. In particular, the linear program searches for a convex combination of these backed-up vectors that pointwise dominates the value vector currently associated with the node. If an improvement is found, the parameters of the convex combination become the new parameters of the node. This interpretation is illustrated in Figure 1, which is adapted from a similar figure from Poupart's dissertation (2005). As Figure 1 shows, the new value vector is parallel to the old vector (i.e., the value of each component is improved by same amount) and it is tangent to the backed-up value function.

**Adding nodes** Eventually, no node can be improved further by solving its corresponding linear program. At this point, BPI is at a local optimum. It can escape such a local optimum by adding one or more nodes to the controller.

The key insight is that when a local optimum is reached, the value vector of each node of the controller is tangent to the backed-up value function at one or more belief states. Moreover, the solution of the linear program that is the dual of the linear program in Table 1 is a belief state that is tangent to the backed-up value function, and so is called the *tangent belief state*. (Since most linear program solvers return both the primal and dual solutions when they solve a linear program, we assume that we get the tangent belief state when we solve the linear program in Table 1, in addition to the value of $\epsilon$ and the action and node transition probabilities.)

To escape a local optimum, it is necessary to improve the value of the tangent belief state. This leads to a method for adding nodes to a controller. Given a tangent belief state $b$, the algorithm considers every belief state $b'$ that can be reached from it in one step (i.e., by taking some action $a$ followed by some observation $z$). For each reachable belief state, a backup is performed (as defined by Equation 3). If the backed-up value is better than the value of the belief state based on the current value function, a deterministic node is added to the controller that has the same action, and, for each observation, the same successor node, that created the improved backed-up value. Often, it is only necessary to add a single node to the controller to escape a local optimum. But because a value vector may be tangent to the backed-up value function for a linear portion of belief space, it may be necessary to add more than one node to escape the local optimum. As we will see, this method for adding nodes to escape local optima is related to the approach that we develop in this paper.

**Two sources of complexity** Most of the computation time of BPI is spent in solving the linear programs that are used to adjust the parameters that specify the action selection probabilities and node transition probabilities of each node of a controller. The size of the linear programs, and thus the complexity of BPI, depends on two things: the size of the controller (which determines the number of variables in the linear program) and the size of the state space (which determines the number of constraints).

|                     |                            || Number of nodes of controller |       |        |        |        |        |
|---------------------|----------------------------|--------|-------|--------|--------|--------|--------|
| Test problem        | Statistic                  || 50     | 100   | 150    | 200    | 250    | 300    |
| Slotted Aloha $\|S\|=30, \|A\|=9, \|Z\|=3$ | total parameters per node  || 1,359  | 2,709 | 4,059  | 5,409  | 6,759  | 8,109  |
|                     | min non-zero parameters    || 6      | 5     | 7      | 7      | 7      | 6      |
|                     | avg non-zero parameters    || 11     | 11    | 11     | 11     | 11     | 11     |
|                     | max non-zero parameters    || 17     | 18    | 20     | 21     | 21     | 21     |
| Tiger grid $\|S\|=36, \|A\|=5, \|Z\|=17$ | total parameters per node  || 4,255  | 8,505 | 12,755 | 17,010 | 21,255 | 25,505 |
|                     | min non-zero parameters    || 45     | 34    | 37     | 33     | 31     | 32     |
|                     | avg non-zero parameters    || 68     | 67    | 68     | 68     | 69     | 68     |
|                     | max non-zero parameters    || 98     | 97    | 97     | 114    | 98     | 114    |
| Hallway $\|S\|=60, \|A\|=5, \|Z\|=21$ | total parameters per node  || 5,255  | 10,505| 15,755 | 21,005 | 26,255 | 31,505 |
|                     | min non-zero parameters    || 28     | 44    | 32     | 35     | 36     | 35     |
|                     | avg non-zero parameters    || 99     | 112   | 105    | 103    | 102    | 100    |
|                     | max non-zero parameters    || 130    | 151   | 158    | 157    | 158    | 158    |
| Hallway2 $\|S\|=92, \|A\|=5, \|Z\|=17$ | total parameters per node  || 4,255  | 8,505 | 12,755 | 17,010 | 21,255 | 25,505 |
|                     | min non-zero parameters    || 46     | 35    | 29     | 39     | 37     | 22     |
|                     | avg non-zero parameters    || 91     | 105   | 112    | 109    | 109    | 114    |
|                     | max non-zero parameters    || 144    | 155   | 163    | 166    | 164    | 167    |

Table 2: Total number of parameters per node of stochastic finite-state controllers found by bounded policy iteration, and minimum, average, and maximum number of parameters with non-zero values, as a function of the size of the controller. (The average is rounded up to the nearest integer.) The four test POMDPs are from (Cassandra, 2004).

In this paper, we focus on the first source of complexity, that is, we focus on improving the complexity of BPI with respect to controller size. (By controller size, we mean not only the number of nodes of the controller, but also the number of actions and observations, since these together determine the number of parameters of a node.) Coping with POMDPs with large state spaces is an orthogonal research issue, and several approaches have been developed that can be combined with BPI. For example, Poupart and Boutilier (2005) describe a technique called *value-directed compression* and report that it allows BPI to solve POMDPs with millions of states. Because the test problems used in this paper have small state spaces, it is important to keep in mind that the techniques developed in the next section for improving the scalability of BPI with respect to controller size can be combined with techniques for coping with large state spaces.

## 3 Sparse bounded policy iteration

In this section, we describe a modified version of BPI that we call *Sparse BPI*. In each iteration, it improves a FSC by the same amount as the original algorithm, but with much improved scalability. To motivate our approach, we begin with a discussion of the sparse structure of stochastic finite-state controllers found by BPI.

### 3.1 Sparse stochastic finite-state controllers

As we have seen, each iteration of BPI solves $|\mathcal{N}|$ linear programs, and each linear program has $|A| + |A||Z||\mathcal{N}|$ variables and $|S| + |A||Z|$ constraints (in addition to the constraints that the variables have non-negative values). Even for small FSCs, the number of variables in the linear programs can be very large, and the fact that the number of variables grows with the number of nodes in the controller significantly limits the scalability of BPI.

If we look at the controllers produced by BPI, however, we find that most of the parameters of each node (i.e., most of the variables in the solutions of the linear program) have values of zero. Table 2 illustrates this vividly for four benchmark POMDPs from (Cassandra, 2004). The overwhelming majority of each node's parameters have zero probabilities. This is despite the fact that *all* of the nodes of these controllers are stochastic. Note that a deterministic node has $1 + |Z|$ non-zero probabilities, one for a choice of action, and $|Z|$ to specify the successor node for each observation. Table 2 shows that the minimum number of non-zero parameters for any node is always greater than this, which indicates that all of the nodes are stochastic. For these problems and many others, BPI is very effective in leveraging the possibility of stochastic nodes to improve a controller without increasing its size. Nevertheless, the actual number of parameters with non-zero probabilities is a very small fraction of the total number of parameters.

Besides the sparsity of the stochastic finite-state controllers found by BPI, an equally important observation about the results in Table 2 is that the number of parameters with non-zero probabilities tends to remain the same as the controller grows in size, whereas the total number of parameters grows significantly. In the following, we develop a modified version of BPI that exploits this sparse structure.

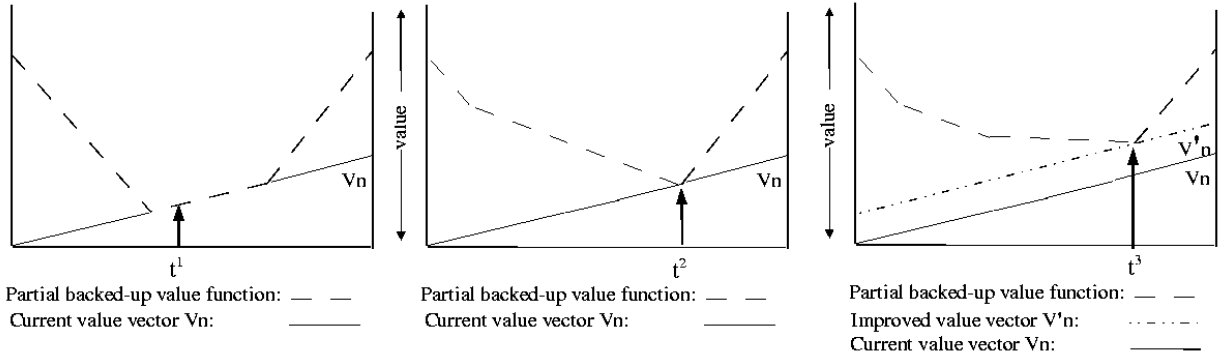

Figure 2: In the left panel, the value vector is tangent to the partial backed-up value function along a linear segment. The center panel shows the result of adding parameters to the sparse linear program that improve the partial backed-up value function at tangent belief state $t^1$. Although this does not improve the value vector, it creates a new tangent belief state $t^2$. When parameters are added to the sparse linear program that improve its backed-up value, the result is an improved value vector with a new tangent belief state $t^3$, as shown in the right panel.

## 3.2 Sparse policy improvement algorithm

We begin by describing the main idea of the algorithm. We have seen that the number of parameters of a node with non-zero probabilities in the solution of the linear program in Table 1 is very small relative to the total number of parameters. Let us call these the *useful parameters*. If we could somehow identify the useful parameters of a node, we could improve a node by solving a linear program that only includes these variables. We call a linear program that only includes some of the parameters of a node a *reduced linear program*. Our approach will be to solve a series of reduced linear programs, where the last one is guaranteed to include all of the useful parameters of the node. This approach will achieve the same result as solving the full linear program of Table 1, but will be more efficient if the reduced linear programs are much smaller.

We next describe an iterative method for identifying a small set of parameters that includes all of the useful parameters. Our method starts with the set of parameters of the node that currently have non-zero probabilities. This guarantees a solution that is at least as good as the current policy for the node. Then we add parameters using a technique that is inspired by the technique BPI uses to add nodes to a controller in order to escape local optima. As shown in Figure 1, the value vector that is created by solving the linear program in Table 1 is tangent to the backed-up value function, and the solution of the dual linear program is a tangent belief state. In the case of a reduced linear program, however, we have a *partial backed-up value function* that does not include vectors corresponding to parameters that are not included in the reduced linear program. (For example, if only one action parameter is included in the reduced linear program, the partial backed-up value function does not include any vectors created by taking a different action.)

If the reduced linear program is missing some useful parameters, it must be possible to improve the value of the tangent belief state $b$ by considering the vectors of the full backed-up value function. So, to identify potentially useful parameters, we perform a backup for the tangent belief state $b$. If the backed-up value, which is defined as

$$\max_{a \in A} \left\{ R(b,a) + \beta \sum_{z \in Z} P(z|b,a) \max_{n \in N} \left( \sum_{s \in S} b_z^a(s) V_n(s) \right) \right\},$$

is greater than the value of the tangent belief state based on the current value vector $V_{n'}$, which is defined as $\sum_{s \in S} b(s) V_{n'}(s)$, we add to our set of parameters the parameters used to create the backed-up value: the best action $a$, and, for each observation $z$, the best successor node $n$. Then we solve another linear program that includes these parameters in addition to the parameters contained in the previous reduced linear program.

If the value vector is tangent to the partial backed-up value function at a single belief state, adding the parameters that improve the backed-up value of this tangent belief state improves the value vector for this node. But since the value vector may be tangent to the partial backed-up value function along a linear segment, adding parameters does not guarantee improvement of the node. In this case, however, it does change the result of the new reduced linear program in an important way. Because adding these parameters to the reduced linear program improves the partial backed-up value function that is searched by the linear program, the value vector is no longer tangent to the partial backed-up value function at the same belief state. In other words, even if the solution of the reduced linear program is not changed by adding these parameters, the solution of its dual linear program changes in one important way: there must be a new tangent belief state. This is illustrated in Figure 2.

This points to an iterative algorithm for improving a node. At each step, the algorithm solves a reduced linear program, and then performs a backup for the tangent belief state. If the backed-up value of the tangent belief state is better than its value based on the current value vector, the parameters that produced the improved backed-up value are added to the reduced linear program. Since the backed-up value of the tangent belief state is improved, it must be the case that at least some of these parameters are not in the current linear program; if they were, the current value vector would not be tangent to the partial backed-up value function at this belief state. The condition for terminating this iterative process of adding parameters to a reduced linear program is based on the following lemma.

**Lemma 1** *The difference between the backed-up value of a tangent belief state and its value based on the current value vector bounds the amount by which the linear program in Table 1 can improve the value vector of a node.*

*Proof:* The linear program in Table 1 implicitly uses the backed-up value function to improve the value vector associated with a node by an amount $\epsilon$ for each state value of the vector; in turn, this improves by $\epsilon$ the value of any belief state based on the value vector. So, if the difference between the backed-up value of any belief state and its value based on the current value vector is $\delta$, then $\epsilon \leq \delta$.□

It follows that if the backed-up value of any belief state is not an improvement of its value based on the current value vector, the linear program cannot improve the value vector.

Based on this lemma, we can prove the following.

**Theorem 1** *This procedure of improving a stochastic node by solving a sequence of reduced linear programs is guaranteed to terminate and the last reduced linear program in the sequence produces the same result as solving the full linear program in Table 1.*

*Proof:* The procedure is guaranteed to terminate because whenever the backed-up value of the tangent belief state is greater than its value based on the value vector created by solving the current reduced linear program, at least one (and no more than $(1 + |Z|)$) parameters will be added to the linear program, and the total number of parameters is finite.

To see that the last reduced linear program solved in this sequence has the same result as solving the full linear program in Table 1, note that the procedure terminates when the difference between the backed-up value of the tangent belief state (for the last reduced linear program) and the value of the tangent belief state based on the best value vector found so far for this node is zero. From Lemma 1, it follows that no further improvement is possible.□

We call this iterative approach to improving the nodes of a stochastic FSC *sparse policy improvement*. The high-level pseudocode is shown in Algorithm 2. Although it can solve multiple linear programs for each linear program solved by the original BPI algorithm, it has an advantage if the number of variables in each of these reduced linear programs is very small compared to the number of variables in the full linear program of Table 1. This is the case whenever the FSCs found by BPI are very sparse.

---

**Algorithm 2** Sparse policy improvement

**for** each node of controller **do**
    Create initial reduced linear program
    {its variables correspond to the parameters of the node that currently have non-zero probabilities}
    $threshold \leftarrow 0$
    **repeat**
        Solve the reduced linear program
        **if** $\epsilon > threshold$ **then**
            Use solution of LP to update parameters of node
            $threshold \leftarrow \epsilon$
        **end if**
        Do backup for tangent belief state
        **if** backed-up value > (current value + $\epsilon$) **then**
            Add variables to linear program that correspond to parameters that produced backed-up value
        **end if**
    **until** no new variables are added to linear program
    increase value vector of node by $\epsilon$
**end for**

---

As mentioned before, this algorithm is inspired by the way the original BPI algorithm adds nodes to a controller. In both cases, the technique for breaking a local optimum at a tangent belief state is to improve the backed-up value function of the tangent belief state. In the original BPI algorithm, the value of the tangent belief state is improved by adding nodes to the controller. In sparse policy improvement, it is improved by adding parameters to a node.

There is also an interesting relationship between this algorithm and column generation methods in linear programming (Bertsekas & Tsitsiklis, 1997). Column generation is a useful strategy for solving linear programs where the number of variables is very large, the number of constraints is relatively small, and most variables have a value of zero in the optimal solution. It begins by considering a small subset of the variables (i.e., the columns) of a problem, and solves a reduced linear program with only these variables. Then it identifies unused variables that can be added to the linear program to improve the result. Use of this strategy requires some way of determining if the current solution is optimal, and if it is not, some way of generating one or more unused variables that can improve the solution. Our algorithm can be viewed as an application of this general strategy to the problem of solving POMDPs using BPI.

|  |  | Number of nodes of controller |  |  |  |  |  |
| Test problem | Algorithm | 50 | 100 | 150 | 200 | 250 | 300 |
|---|---|---|---|---|---|---|---|
| Slotted Aloha | BPI | 202 | 415 | 684 | 871 | 1,085 | 1,620 |
| $|S|=30, |A|=9, |Z|=3$ | Sparse-BPI | 16 | 19 | 19 | 20 | 21 | 23 |
| Tiger grid | BPI | 1,365 | 3,447 | 5,822 | 9,461 | 12,358 | 15,770 |
| $|S|=36, |A|=5, |Z|=17$ | Sparse-BPI | 189 | 212 | 335 | 174 | 207 | 274 |
| Hallway | BPI | 3,900 | 10,180 | 17,340 | 24,485 | 27,428 | 32,973 |
| $|S|=60, |A|=5, |Z|=21$ | Sparse-BPI | 1,215 | 935 | 1,110 | 973 | 1,094 | 1,267 |
| Hallway2 | BPI | 7,760 | 17,729 | 29,809 | 42,905 | 53,903 | 68,898 |
| $|S|=92, |A|=5, |Z|=17$ | Sparse-BPI | 2,668 | 4,128 | 3,323 | 4,078 | 3,513 | 3,388 |

Table 3: Average time (in CPU milliseconds) for improving a single node of a finite-state controller, as a function of the size of the controller, for four benchmark POMDPs.

## 4 Experimental results

We implemented sparse bounded policy iteration and tested it on several benchmark POMDPs. Table 3 shows results for the four POMDPs considered earlier in Table 2. The experiments ran on a 3.0 GHz processor using a linear program solver in CPLEX version 9.0. We solved each POMDP beginning with an initial controller with $|A|$ nodes, and adding up to five new nodes each time the policy improvement step reached a local optimum. Table 3 reports the average running time to improve a single node because this dominates overall computation time, and also because it measures performance independently of design decisions such as how and when to add nodes to a controller.

Table 3 shows that Sparse BPI is much more efficient than the original BPI algorithm in improving sparse stochastic FSCs. Even for relatively small controllers of 300 nodes, Sparse BPI runs between 20 and 80 times faster than BPI in solving these particular POMDPs. More importantly, its relative advantage grows with the size of the controller. The size of the reduced linear programs solved by Sparse BPI remains about the same as the controller grows in size; as a result, the time it takes to improve a single node of the controller remains relatively constant as the size of the controller grows, in contrast to BPI.

An interesting difference between Sparse BPI and BPI is that the running time of an iteration of Sparse BPI depends on how much improvement of the controller is possible. If the nodes of a controller are already at a local optimum, Sparse BPI often needs to solve only a couple reduced linear programs per node in order to determine that further improvement is not possible. In this case, an iteration of Sparse BPI terminates relatively quickly. But if much improvement of the FSC is possible, Sparse BPI often needs to solve ten or twenty or even more reduced linear programs for some nodes in order to add all of the parameters that are needed to maximize improvement of the node. To obtain reliable running times for Sparse-BPI, the results in Table 3 are averaged over several iterations of Sparse-BPI.

Table 3 shows that a larger state space slows both algorithms, but it slows the sparse algorithm more. This is because the sparse algorithm solves multiple linear programs for each linear program solved by the original algorithm, and more constraints makes these more difficult to solve.

The running time of Sparse BPI could be reduced further by finding a way to reduce the number of iterations it takes to improve a node. For the *hallway2* problem, for example, the sparse algorithm can take ten or more iterations to improve a node – occasionally, as many as thirty or forty. Each iteration requires solving a reduced linear program. But in fact, it is only necessary to solve one linear program to change the parameters of a node. The other linear programs in the sequence of reduced linear programs solved by the sparse algorithm are used to identify a sequence of tangent belief states; for each tangent belief state, a backup is performed in order to identify parameters to add to the stochastic node. It seems possible to identify a set of belief states for which to perform backups without needing to solve a linear program to identify each one.

Point-based methods solve POMDPs by performing a backup for each of a finite set of belief states (Pineau, Gordon, & Thrun, 2003). Recent work shows that this approach can be combined with policy iteration algorithms that represent a policy as a finite-state controller (Ji, Parr, Li, Liao, & Carin, 2007). An interesting direction of research is to use point-based backups to identify parameters that can be added to nodes; after many backups are performed, a reduced linear program could be solved for each node in order to improve its action selection and node transition probabilities – integrating point-based methods with policy iteration for stochastic finite-state controllers.

Another possible way to reduce the number of iterations is early termination. Instead of continuing to add parameters until the difference between the backed-up value of the tangent belief state and its value based on the current value vector is zero, Sparse-BPI could stop adding parameters as soon as the difference is small enough to demonstrate that only a small amount of further improvement is possible.

## 5 Conclusion

We have presented a modified bounded policy iteration algorithm for POMDPs called sparse bounded policy iteration. The new algorithm exploits the sparse structure of stochastic finite-state controllers found by bounded policy iteration. Each iteration of the algorithm produces the identical improvement of a controller that an iteration of the original bounded policy iteration algorithm produces, but with improved scalability. Whereas the time it takes for the original algorithm to improve a single node grows with the size of the controller, the time it takes for the new algorithm to improve a single node is typically independent of the size of the controller. This makes it practical to use bounded policy iteration to find larger controllers for POMDPs.

### Acknowledgements

This work was supported in part by the Air Force Office of Scientific Research under Award No. 05-003202A05.